\documentclass[10pt,twocolumn,letterpaper]{article}

\usepackage{cvpr}              

\usepackage{graphicx}
\usepackage{amsmath}
\usepackage{amssymb}
\usepackage{booktabs}	
\usepackage{bm}

\usepackage[pagebackref,breaklinks,colorlinks]{hyperref}

\usepackage[capitalize]{cleveref}
\crefname{section}{Sec.}{Secs.}
\Crefname{section}{Section}{Sections}
\Crefname{table}{Table}{Tables}
\crefname{table}{Tab.}{Tabs.}

\usepackage{marvosym}
\usepackage{balance}
\usepackage{arydshln}
\usepackage{multirow}
\usepackage{makecell}

\newcommand{\tablestyle}[2]{\setlength{\tabcolsep}{#1}\renewcommand{\arraystretch}{#2}\centering\footnotesize}
\newlength\savewidth\newcommand\shline{\noalign{\global\savewidth\arrayrulewidth
		\global\arrayrulewidth 1pt}\hline\noalign{\global\arrayrulewidth\savewidth}}
\usepackage[table]{xcolor}

\def\cvprPaperID{6795} 
\def\confName{CVPR}
\def\confYear{2022}

\newcommand*{\affaddr}[1]{#1} 
\newcommand*{\affmark}[1][*]{\textsuperscript{#1}}

\makeatletter
\newcommand{\printfnsymbol}[1]{\textsuperscript{\@fnsymbol{#1}}}
\makeatother

\makeatletter
\def\thanks#1{\protected@xdef\@thanks{\@thanks
        \protect\footnotetext{#1}}}
\makeatother

\begin{document}

\title{Multimodal Token Fusion for Vision Transformers}

\author{
	Yikai Wang\affmark[1]\quad Xinghao Chen\affmark[2]\quad Lele Cao\affmark[1]\quad Wenbing Huang\affmark[3]\quad\vspace{0.05in}Fuchun Sun\affmark[1]\textsuperscript{\Letter}$\thanks{\textsuperscript{\Letter}~Corresponding author: Fuchun Sun.}$\quad Yunhe Wang\affmark[2]\\
	\affaddr{\affmark[1]Beijing National Research Center for Information Science and Technology$\,$(BNRist),\\ State Key Lab on Intelligent Technology and Systems,\\ Department of Computer Science and Technology, Tsinghua University} \\\affaddr{\affmark[2]Huawei Noah’s Ark Lab}\quad
	\affaddr{\affmark[3]Institute for AI Industry Research (AIR), Tsinghua University}\\
	\tt\small{wangyk17@mails.tsinghua.edu.cn, xinghao.chen@huawei.com, caolele@gmail.com, }\\
	\tt\small{hwenbing@126.com, fuchuns@tsinghua.edu.cn, yunhe.wang@huawei.com}\\
}

\maketitle

\begin{abstract}

Many adaptations of transformers have emerged to address the single-modal vision tasks, where self-attention modules are stacked to handle input sources like images. Intuitively, feeding multiple modalities of data to vision transformers could improve the performance, yet the inner-modal attentive weights may also be diluted, which could thus undermine the final performance. In this paper, we propose a multimodal token fusion method (TokenFusion), tailored for transformer-based vision tasks. To effectively fuse multiple modalities, TokenFusion dynamically detects uninformative tokens and substitutes these tokens with projected and aggregated inter-modal features. Residual positional alignment is also adopted to enable explicit utilization of the inter-modal alignments after fusion. The design of TokenFusion allows the transformer to learn correlations among multimodal features, while the single-modal transformer architecture remains largely intact. Extensive experiments are conducted on a variety of homogeneous and heterogeneous modalities and demonstrate that TokenFusion surpasses state-of-the-art methods in three typical vision tasks: multimodal image-to-image translation, RGB-depth semantic segmentation, and 3D object detection with point cloud and images. Our code is available at \url{https://github.com/yikaiw/TokenFusion}. 
\vskip-0.01in
\end{abstract}

\section{Introduction}
\label{sec:intro}
Transformer is initially widely studied in the natural language community as a non-recurrent sequence model~\cite{vaswani2017attention} and it is soon extended to benefit vision-language tasks. Recently, numerous studies have further adopted transformers for computer vision tasks with well-adapted architectures and optimization schedules. As a result, vision transformer variants have shown great potential in many single-modal vision tasks, such as classification~\cite{dosovitskiy2020image,liu2021swin}, segmentation~\cite{xie2021segformer,SETR}, detection~\cite{carion2020end,fang2021you,zhu2020deformable,liu2021group}, image generation~\cite{jiang2021transgan}. 

Yet up until the date of this work, the attempt of extending vision transformers to handle multimodal data remains scarce. When multimodal data with complicated alignment relations are introduced, it poses great challenges in designing the fusion scheme for model architectures. 
The key question to answer is how and where the interaction of features from different modalities should take place. 
There have been a few methods for transformer-based vision-language fusion, \eg, VL-BERT~\cite{su2019vl} and ViLT~\cite{kim2021vilt}. In these methods, vision and language tokens are directly concatenated  before each transformer layer, making the overall architecture very similar to the original transformer. Such fusion is usually alignment-agnostic, which indicates the inter-modal alignments are not explicitly utilized.
We also try to apply similar fusion methods on multimodal vision tasks (Sec.~\ref{sec:experiments}). Unfortunately, this intuitive transformer fusion cannot bring promising gains or may even result in worse performance than the single-modal counterpart, which is mainly due to the fact that the inter-modal interaction is not fully exploited.
There are also several attempts for fusing multiple vision modalities. For example, TransFuser~\cite{prakash2021multi} leverages  transformer modules to connect CNN backbones of  images and LiDAR points. Different from exising trials, our work aims to seek an effective and general method to combine multiple single-modal transformers while inserting inter-modal alignments into the models.

This work  benefits the  learning process by  multimodal data while   leveraging inter-modal alignments. Such alignments are naturally available in many vision tasks, \eg, with camera intrinsics/extrinsics, world-space points could be projected and correspond to pixels on the camera plane.
Unlike the alignment-agnostic fusion (\cref{subsec:intuitive_fusion}), the alignment-aware fusion explicitly involves the alignment relations of different modalities. Yet, since inter-modal projections are introduced to the transformer, alignment-aware fusion may greatly alter the original model structure and data flow, which potentially undermines the success of single-modal architecture designs or learned attention during pretraining. Thus, one may have to determine the ``correct'' layers/tokens/channels for multimodal projection and fusion, and also re-design the architecture or re-tune optimization settings for the new model.
To avoid dealing with these challenging matters and inherit the majority of the original single-modal design, 
we propose multimodal token fusion, termed \textbf{TokenFusion}, which adaptively and effectively fuses multiple single-modal transformers. 

The basic idea of our TokenFusion is to prune multiple single-modal transformers and then re-utilize pruned units for multimodal fusion. 
We apply individual pruning to each single-modal transformer and each pruned unit is substituted by projected alignment features from other modalities. 
This fusion scheme is assumed to have a limited impact on the original single-modal transformers, as it maintains the relative attention relations of the {important} units.  TokenFusion also turns out to be superior in allowing multimodal transformers to inherit the parameters from single-modal pretraining, \eg, on ImageNet.

To demonstrate the advantage of the proposed method, we consider extensive tasks including multimodal image translation, RGB-depth semantic segmentation, and 3D object detection based on images and point clouds, covering up to four public datasets and seven different modalities.  TokenFusion obtains state-of-the-art performance on these extensive tasks, demonstrating its great effectiveness and generality. Specifically,  TokenFusion achieves 64.9\% and 70.8\% mAP@0.25 for 3D object detection on the challenging SUN RGB-D and ScanNetV2 benchmarks, respectively.

\section{Related Work}
\label{sec:related_work}

\textbf{Transformers in computer vision.} Transformer is originally designed for NLP research fields~\cite{vaswani2017attention}, which stacking multi-head self-attention and feed-forward MLP layers to capture the long-term correlation between words. Recently, vision transformer (ViT)~\cite{dosovitskiy2020image} reveals the great potential of transformer-based models in large-scale image classification. As a result, transformer has soon achieved profound impacts in many other computer vision tasks such as segmentation~\cite{xie2021segformer,SETR}, detection~\cite{carion2020end,fang2021you,zhu2020deformable,liu2021group}, image generation~\cite{jiang2021transgan}, video processing~\cite{DBLP:journals/corr/abs-2109-02974}, etc.

\textbf{Fusion for vision transformers.}
Deep  fusion with multimodal data has been an essential topic which potentially boosts the performance by leveraging multiple sources of inputs, and it  may also unleash the power of transformers further. Yet it is challenging to combine multiple off-the-rack single transformers  while guaranteeing that such combination will not impact their elaborate singe-modal designs. \cite{DBLP:journals/corr/abs-2107-02191} and \cite{DBLP:journals/corr/abs-2109-02974} process consecutive video frames with transformers for spatial-temporal alignments and capturing fine-grained patterns by correlating multiple frames. Regarding multimodal data, \cite{prakash2021multi,vs2021image} utilize the dynamic property of transformer modules to combine CNN backbones for fusing infrared/visible images or LiDAR points. ~\cite{fu2021ppt} extends the coarse-to-fine experience from CNN fusion methods to transformers for image processing tasks. \cite{hu2021fusformer} adopts transformers to combine hyperspectral images by the simple feature concatenation.  \cite{nagrani2021attention} inserts intermediate tokens between image patches and audio spectrogram patches as bottlenecks to implicitly learn inter-modal alignments.
These works, however, differ from ours since  we would like to build a general fusion pipeline for combing off-the-rack vision transformers without the need of re-designing their structures or re-tuning their optimization settings, while explicitly leveraging inter-modal alignment relations.

\section{Methodology}
\label{sec:methodology}
This part intends to provide a full landscape of the proposed methodology. We first introduce two na\"ive multimodal fusion methods for vision transformers in \cref{subsec:intuitive_fusion}. Given the limitations of both intuitive methods, we then propose multimodal token fusion in \cref{subsec:mix_transformers}. We elaborate the fusion designs for both homogeneous and heterogeneous  modalities to evaluate the effectiveness and generality of our method in \cref{subsec:homogeneous-modalities} and \cref{subsec:heterogeneous-modalities}, respectively.

\subsection{Basic Fusion for Vision Transformers}
\label{subsec:intuitive_fusion}
Suppose we have the $i$-th input data $\bm{x}^{(i)}$ that contains $M$ modalities: $\bm{x}^{(i)}=\{\bm{x}_m^{(i)}\in\mathbb{R}^{N\times C}\}_{m=1}^M$, 
where $N$ and $C$ denote the number of tokens and input channels respectively. 
For simplicity, we will omit the subscript $^{(i)}$ in the upcoming sections. 
The goal of deep multimodal fusion is to determine a multi-layer model $f(\bm{x})$, 
and its output is expected to  close to the target $\bm{y}$ as much as possible. 
Specifically in this work, $f(\bm{x})$ is approximated by a transformer-based network architecture. Suppose the model contains $L$ layers in total, we represent the input token feature of the $l$-th layer ($l=1,\ldots,L$) as $\bm{e}^l=\{\bm{e}^l_m\in\mathbb{R}^{N\times C'}\}_{m=1}^M$, 
where $C'$ denotes the number of feature channels of the layer in scope. 
Initially, $\bm{e}_m^1$ is obtained using a linear projection of $\bm{x}_m$, which is a widely adopted approach to vectorize the input tokens (\emph{e.g.} image patches), so that the first transformer layer can accept tokens as input.

We use different transformers for input modalities and denote $f_m(\bm{x})=\bm{e}_m^{L+1}$ as the final prediction of the $m$-th transformer.
Given the token feature $\bm{e}_m^{l}$ of the $m$-th modality, the $l$-th layer computes \vskip-0.2in
\begin{equation}
\label{eq:token-feature}
\hat{\bm{e}}_m^l=\text{MSA}\big(\text{LN}(\bm{e}_m^{l})\big),\; \bm{e}_m^{l+1}=\text{MLP}\big(\text{LN}(\hat{\bm{e}}_m^l)\big),
\end{equation}
where $\text{MSA}$, $\text{MLP}$, and $\text{LN}$ denote the multi-head self-attention, multi-layer perception, and layer normalization, receptively. 
$\hat{\bm{e}}^l_m$ represents the output of MSA. 

During multimodal fusion for vision tasks, the alignment relations of different modalities may be explicitly available. For example, pixel positions are often used to determine the image-depth correlation; and camera intrinsics/extrinsics are important in projecting 3D points to images. Based on the involvement of alignment information, we consider two kinds of transformer fusion methods as below.

\textbf{Alignment-agnostic fusion} does not explicitly use the alignment relations among modalities. 
It expects the alignment may be implicitly learned from large amount of data. 
A common method of the alignment-agnostic fusion is to directly concatenate multimodal input tokens, which is widely applied in vision-language models. Similarly, the input feature $\bm{e}_l$ for the $l$-th layer is also the token-wise concatenation of different modalities. Although the alignment-agnostic fusion is simple and may have minimal modification to the original transformer model, 
it is hard to directly benefit from the known multimodal alignment relations.

\textbf{Alignment-aware fusion}
\label{sec:alignment-aware-fusion}  explicitly utilizes  inter-modal alignments. 
For instance, this can be achieved by selecting tokens that correspond to the same pixel or 3D coordinate.
Suppose $\bm{x}_m[n]$ is the $n$-th token of the $m$-th modality input $\bm{x}_m$, where $n=1,\cdots,N_m$. 
We define the ``token projection'' from the $m$-th modality to the $m'$-th modality as 
\begin{equation}
\label{eq:token-projection}
\mathrm{Proj}^\text{T}_{m'}(\bm{x}_{m}[n_{m}])=h(\bm{x}_{m'}[n_{m'}]),
\end{equation}
where $h$ could simply be an identity function (for homogeneous modalities) or a shallow multi-layer perception (for heterogeneous modalities). 
And when considering the entire $N$ tokens, we can conveniently define the ``modality projection'' as the concatenation of token projections:\vskip-0.2in
\begin{equation}
\label{eq:modality-projection}
{\mathrm{Proj}}^\text{M}_{m'}(\bm{x}_{m})=\big[\mathrm{Proj}^\text{T}_{m'}(\bm{x}_{m}[1]); \cdots; \mathrm{Proj}^\text{T}_{m'}(\bm{x}_{m}[N])\big].
\end{equation}

\cref{eq:modality-projection} only depicts the fusion strategy on the input side. We can also perform middle-layer or multi-layer fusion across different modality-specific models, by projecting and aggregating feature embeddings $\bm{e}_m$ which possibly enables more diversified and accurate feature interactions. 
However, with the growing complexity of transformer-based models, searching for optimal fusion strategies (\emph{e.g.} layers and tokens to apply projection and aggregation) for merely two modalities (\emph{e.g.} 2D and 3D detection transformers) can grow into an extremely hard problem to solve.  To tackle this issue, we propose multimodal token fusion in \cref{subsec:mix_transformers}.

\subsection{Multimodal Token Fusion}
\label{subsec:mix_transformers}

As described in \cref{sec:intro}, multimodal token fusion (TokenFusion) first prunes single-modal transformers and further re-utilizes the pruned units for  fusion. In this way, the informative units of original single-modal transformers are assumed to be preserved to a large extent, while multimodal interactions could be  involved for boosting performance. 

As previously shown in~\cite{DBLP:journals/corr/abs-2106-02034}, tokens of vision transformers could be pruned in a hierarchical manner while maintaining the performance.
Similarly, we can select less informative tokens
by adopting a scoring function $s^l(\bm{e}^{l})=\text{MLP}(\bm{e}^{l})\in[0,1]^N$, 
which dynamically predicts the importance of tokens for the $l$-th layer and the $m$-th modality. 
To enable the back propagation on $s^l(\bm{e}^{l})$, we re-formulate the MSA output $\hat{\bm{e}}_m^l$ in \cref{eq:token-feature} as 
\begin{equation}
\label{eq:msa-output-score}
\hat{\bm{e}}_m^l=\text{MSA}\big(\text{LN}(\bm{e}_m^{l})\cdot s^l(\bm{e}_m^{l})\big).
\end{equation}

We use $\mathcal{L}_m$ to denote the task-specific loss for the $m$-th modality. To prune uninformative tokens, we further add a token-wise pruning loss (an $l_1$-norm) on $s^l(\bm{e}_m^{l})$. Thus the overall loss function for optimization is derived as \vskip-0.1in
\begin{equation}
\label{eq:overall-loss}
\mathcal{L}=\sum_{m=1}^M\Big(\mathcal{L}_m+\lambda\sum_{l=1}^L\big|s^l(\bm{e}_m^{l})\big|\Big),
\end{equation}
where $\lambda$ is a hyper-parameter for balancing different losses. 

For the feature $\bm{e}_m^l\in\mathbb{R}^{N\times C'}$, token-wise pruning dynamically detects unimportant tokens from all $N$ tokens. Mutating unimportant tokens or substituting them with other embeddings are expected to have limited impacts on other informative tokens. We thus propose a token fusion process for multimodal transformers, which substitute unimportant tokens with their token projections (defined in \cref{sec:alignment-aware-fusion}) from other modalities. 
Since the pruning process is dynamic, {\emph{i.e.}}, conditioned on the input features, the fusion process is also dynamic. 
This process performs token substitution before each transformer layer, thus the input feature of the $l$-th layer, \emph{i.e.}, $\bm{e}_m^l$, is re-formulated as \vskip-0.2in
\begin{align}
\label{eq:mix-process}
\bm{e}_m^l&=\bm{e}_m^l\odot\mathbb{I}_{s^l(\bm{e}_m^{l})\ge\theta}
+{\mathrm{Proj}}^\text{M}_{m'}(\bm{e}_m^l)\odot\mathbb{I}_{s^l(\bm{e}_m^{l})<\theta},
\end{align}
where $\mathbb{I}$ is an indicator asserting the subscript condition, therefore it outputs a mask tensor $\in\{0,1\}^N$; the parameter $\theta$ is a small threshold (we adopt $10^{-2}$ in our experiments); and the operator $\odot$ resents the element-wise multiplication.

In \cref{eq:mix-process}, if there are only two modalities as input, $m'$ will simply be the other modality other than $m$. 
With more than two modalities, we pre-allocate the tokens into $M-1$ parts, each of which is bound with one of the other modalities than itself. 
More details of this pre-allocation will be described in \cref{subsec:homogeneous-modalities}.

\subsection{Residual Positional Alignment} 
\label{subsec:rpa}
\begin{figure}[t!]
\centering
\hskip0.07in
\includegraphics[scale=0.42]{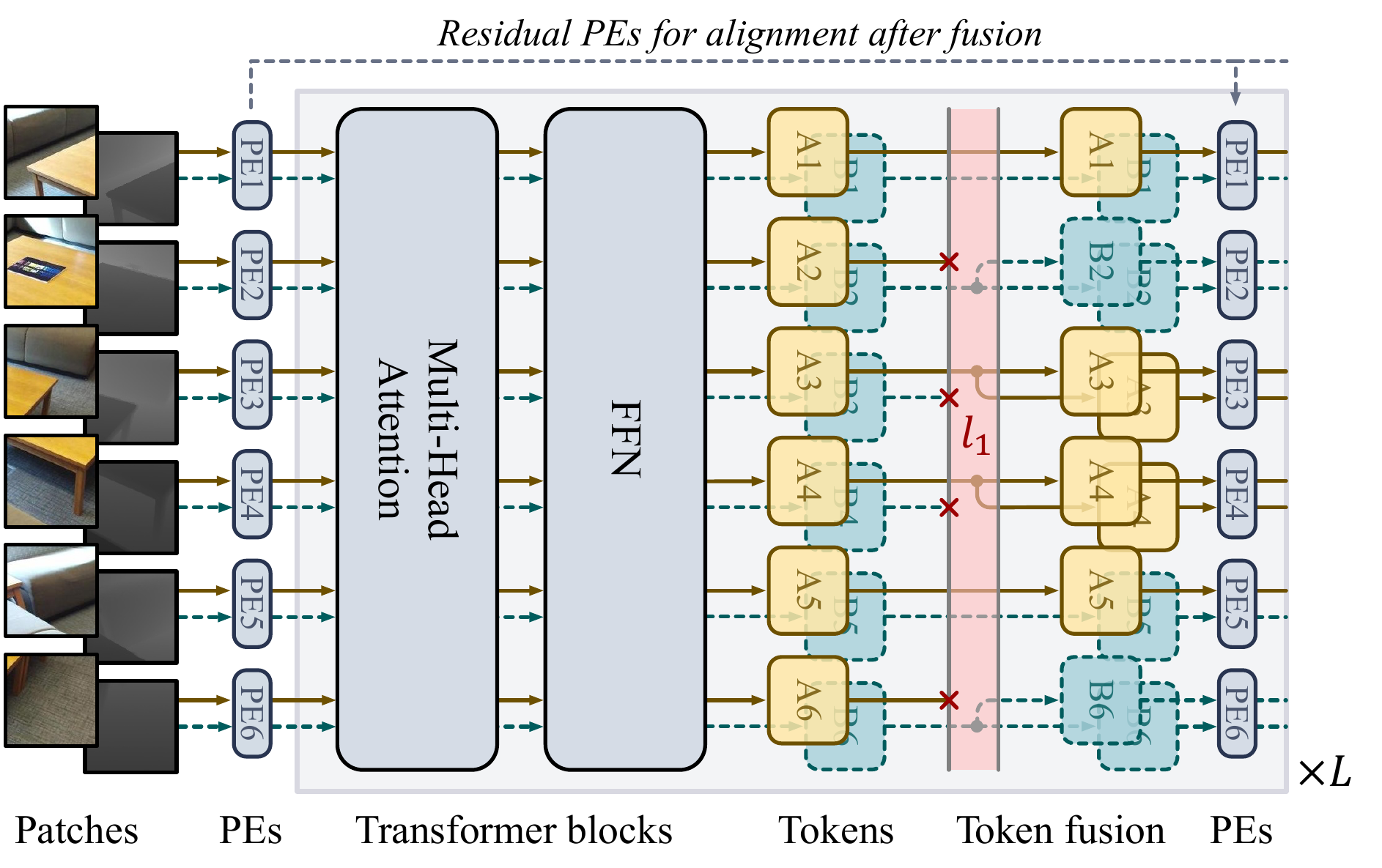}
\caption[]{Framework of TokenFusion for homogeneous modalities with RGB and depth as an example. Both modalities are sent to a shared transformer  with also shared positional embeddings. }
\label{pic:framework}
\vskip-0.1in
\end{figure}

\begin{figure*}[t!]
\centering
\includegraphics[scale=0.42]{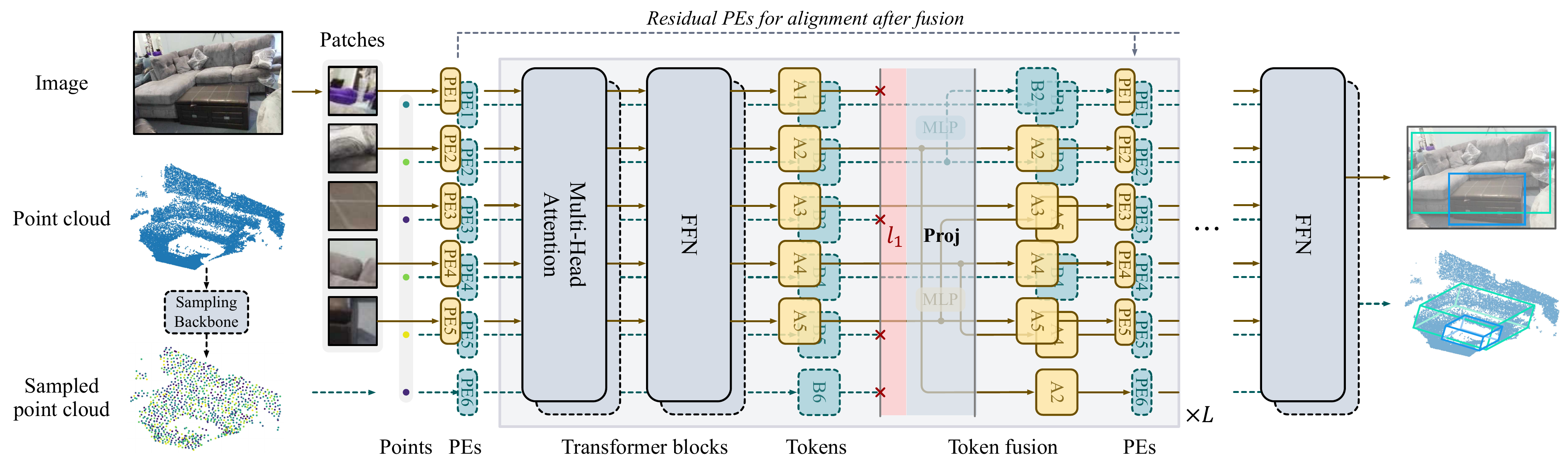}
\caption[]{Framework of TokenFusion for heterogeneous modalities with point clouds and images. Both modalities are sent to individual transformer modules with also individual positional embeddings. Additional inter-modal projections (Proj) are needed which is different from the fusion for homogeneous modalities.
}
\label{pic:framework-hete}
\vskip-0.1in
\end{figure*}

Directly substituting tokens will risk completely undermining their original positional information. 
Hence, the model can still be ignorant of the alignment of the projected features from another modality. 
To mitigate this problem, we  adopt Residual Positional Alignment (RPA)  that leverages Positional Embeddings (PEs) for the multimodal alignment. As depicted in \cref{pic:framework} and \cref{pic:framework-hete} which will be detailed later, the key idea of RPA lies in injecting equivalent PEs to subsequent layers. 
Moreover, the back propagation of PEs stops after the first layer, which means only the gradients of PEs  at the first layer are retained while for the rest of the layers are frozen throughout the training.  In this way, PEs serve a purpose of aligning multimodal tokens despite the substitution status of the original token.
In summary, even if a token is substituted, we still reserve its original PEs that are added to the projected feature from another modality.

\subsection{Homogeneous Modalities} 
\label{subsec:homogeneous-modalities}

In the common setup of either a generation task (multimodal image-to-image translation) or a regression task (RGB-depth semantic segmentation), the homogeneous vision modalities $\bm{x}_1,\bm{x}_2,\cdots, \bm{x}_M$ are typically aligned with pixels, such that the pixels located at the same position in RGB or depth input should share the same label. 
We also expect that such property allows the transformer-based models to benefit from joint learning. 
Hence, we adopt shared parameters in both MSA and MLP layers for different modalities; yet rely on modality-specific layer normalizations to uncouple the normalization process, 
since different modalities may vary drastically in their statistical means and variances by nature. 
In this scenario, we simply set function $h$ in \cref{eq:mix-process} as an identity function, 
and we also let $n_{m'}=n_m$, which means we always substitute each pruned token with the token sharing the same position.  

An overall illustration of TokenFusion for fusing homogeneous modalities is depicted in \cref{pic:framework}. 
Regarding two input modalities, we adopt bi-directional projection and apply token-wise pruning on both modalities respectively. 
Then the token substitution process is performed according to \cref{eq:mix-process}. 
When there are $M>2$ modalities, we also apply the token-wise pruning on all modalities with an additional pre-allocation strategy that selects $m'$ in based on $m$ according to \cref{eq:mix-process}. 
To be specific, for the $m$-th modality, we randomly pre-allocate $N$ tokens into $M-1$ groups with equal group sizes. 
This pre-allocation is carried out prior to the commence of training procedure, and the obtained groups will be fixed throughout the training.
We denote the group allocation as $\bm{a}_{m'}(m)\in\{0,1\}^N$, where $\bm{a}_{m'}(m)[n]=1$ indicates that if the $n$-th token of the $m$-th modaltity is pruned, it will be  substituted by the corresponding token of the $m'$-th modality, otherwise $\bm{a}_{m'}(m)[n]=0$.
Having obtained the pre-allocation strategy for $M>2$ modalties, \cref{eq:mix-process} can be further developed into a more specific form:
\begin{align}
\label{eq:mix-process-allocated}
\nonumber
\bm{e}_m^l&=\bm{e}_m^l\odot\mathbb{I}_{s^l(\bm{e}_m^{l})\ge\theta}\\
&+\sum_{\substack{m'=1\\m'\ne m}}^M{\bm{a}_{m'}(m)\odot\mathrm{Proj}}^\text{M}_{m'}(\bm{e}_m^l)\odot\mathbb{I}_{s^l(\bm{e}_m^{l})<\theta}.
\end{align}

\subsection{Heterogeneous Modalities} 
\label{subsec:heterogeneous-modalities}
In this section, we further explore how TokenFusion handles heterogeneous modalities, in which input modalities exhibit quite different data formats and large structural discrepancies, ~\eg, different number of layers or embedding dimensions for the transformer architectures. 
A concrete example would be to learn 3D object detection (based on point cloud) and 2D object detection (based on images) simultaneously with different transformers. 
Although there are already specific transformer-based models designed for 3D or 2D object detection respectively, there still lacks a fast and effective method to combine these models and tasks.

An overall structure of  TokenFusion for fusing heterogeneous modalities is depicted in ~\cref{pic:framework-hete}. 
Different from the homogeneous case, we approximate the token projection function $h$ in \cref{eq:token-projection} with a shallow multi-layer perception (MLP), since transformers for these heterogeneous modalities may have different hidden embedding dimensions. 
For the case of 3D object detection with 3D point cloud and 2D image, we project each point to the corresponding image based on camera intrinsics and extrinsics. 
Likewise, we also project 3D object labels to the images for obtaining the corresponding 2D object labels. 
We train two standalone transformers with unshared parameters in an end-to-end manner. 
Regarding the 3D object detection with point cloud as input, we follow the architecture used in Group-Free~\cite{liu2021group}, where $N_\text{point}$ sampled {seed points} and $K_\text{point}$ learned {proposal points} are considered as input tokens, which are sent to the transformer for predicting $K_\text{point}$ 3D bounding boxes and object categories. For the 2D object detection with images as input, we follow the  framework in YOLOS~\cite{fang2021you} which  sends $N_\text{img}$ image patches and $K_\text{img}$ object queries to the transformer to predict $K_\text{img}$ 2D bounding boxes together with their associated object categories.

The inter-modal projection maps seed points to image patches, ~\ie, an $N_\text{point}$-to-$N_\text{img}$ mapping. 
Specifically, the token-wise pruning is applied on the $N_\text{point}$ seed point tokens. 
Once a certain token obtains a low importance score, we project the 3D coordinate of this token to a 2D pixel on the corresponding image input. 
It is now viable to locate the specific image patch based on the 2D pixel. 
Suppose this projection obtains the $n_\text{img}$-th image patch based on the $n_\text{point}$-th seed point which is pruned. 
We substitute $m$ and $m'$ in \cref{eq:token-projection} with the subscripts ``$\text{point}$'' and ``$\text{img}$'' respectively, ~\ie, $\mathrm{Proj}^\text{T}_\text{img}(\bm{x}_\text{point}[n_\text{point}])=h(\bm{x}_\text{img}[n_\text{img}])$. 
Thus the relation between $n_\text{point}$ and $n_\text{img}$ captured by the token projection satisfies
\begin{align}
\label{eq:projection}
\big[u,\,v,\,z\big]^\top&=\bm{\mathrm{K}}\cdot\bm{\mathrm{R_t}}\cdot\big[x_{n_\text{point}},\,y_{n_\text{point}},\,z_{n_\text{point}},\,1\big]^\top,\\
n_\text{img}&=\Big\lfloor\frac{\lfloor v/z\rfloor}{P}\Big\rfloor\times \Big\lfloor\frac{W}{P}\Big\rfloor+\Big\lfloor\frac{\lfloor u/ z\rfloor}{P}\Big\rfloor,
\end{align}
where $\bm{\mathrm{K}}\in\mathbb{R}^{4\times 4}$ and $\bm{\mathrm{R_t}}\in\mathbb{R}^{4\times 4}$ are camera intrinsic and extrinsic matrices, respectively; $[x_{n_\text{point}},\,y_{n_\text{point}},\,z_{n_\text{point}}]$ denotes the 3D coordinate of the $n_\text{point}$-th point; $u,v,z$ are temporary variables with $\big[\lfloor u/z\rfloor,\,\lfloor v/z\rfloor\big]$ actually being the projected pixel coordinate of the image; $P$ is the patch size of the vision transformer and $W$ denotes the image width.

\section{Experiments}
\label{sec:experiments}

\begin{figure*}[t!]
\centering\vskip-0.1in
\includegraphics[scale=0.16]{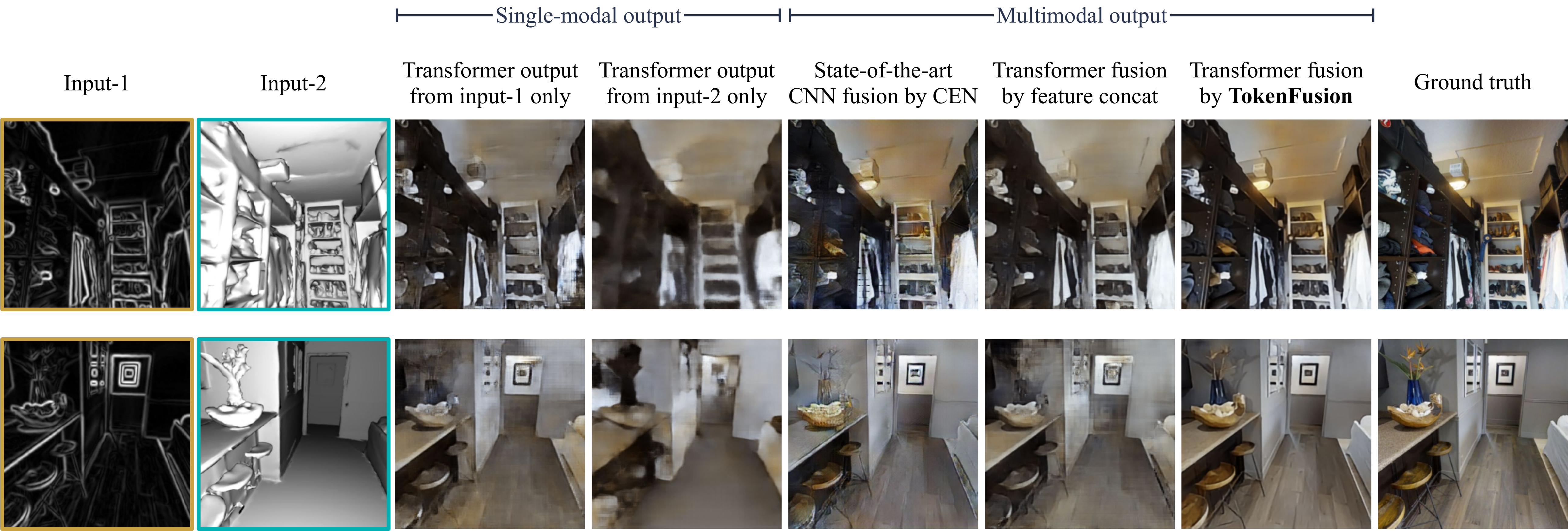}
\caption[]{Comparison on the \emph{validation} data split for image-to-image translation (Texture+Shade$\to$RGB). The resolution of all input/output images is $256\times256$. The third/forth column is predicted by the single modality, and the following three columns are predicted by CEN \cite{DBLP:conf/nips/WangHSXRH20}, the intuitive transformer fusion by feature concatenation, and our TokenFusion, respectively.  {Best view in color and zoom in.}}
\label{pic:pix2pix-compare}\vskip-0.06in
\end{figure*}

To evaluate the effectiveness of the proposed TokenFusion,
we conduct comprehensive experiments towards both homogeneous and heterogeneous modalities with state-of-the-art (SOTA) methods.
Experiments are conducted on totally seven different modalities and four application scenarios, implemented with PyTorch~\cite{conf/nips/PaszkeGMLBCKLGA19} and MindSpore~\cite{mindspore}.

\subsection{Multimodal Image-to-Image Translation} 
\label{subsec:exp-img2img}

The task of multimodal image-to-image translation aims at generating a target image modality based on different image modalities as input (\emph{e.g.} Normal+Depth$\to$RGB). We evaluate TokenFusion in this task using the Taskonomy~\cite{conf/cvpr/ZamirSSGMS18} dataset,
which is a large-scale indoor scene dataset containing about 4 million indoor images captured from 600 buildings. 
Taskonomy provides over 10 multimodal representations in addition to each RGB image, such as depth (euclidean or z-buffering), normal, shade, texture, edge, principal curvature, etc. 
The resolution of each representation is $512\times512$. 
To facilitate comparison with the existing fusion methods, we adopt the same sampling strategy as ~\cite{DBLP:conf/nips/WangHSXRH20}, resulting in 1,000 high-quality multimodal images for training, and 500 for validation. 

Our implementation contains two transformers as the generator and discriminator respectively. We provide configuration details in our supplementary materials. The resolution of the generator/discriminator input or the generator prediction is  $256\times256$. We adopt two kinds of architecture settings, the tiny (Ti) version with $10$ layers and the small (S) version with $20$ layers, and both settings are only different in layer numbers. The learning rates of both transformers are set to $2\times 10 ^{-4}$. We adopt overlapped patches in both transformers inspired by~\cite{xie2021segformer}. 

In our experiments for this task, we adopt shared transformers for all input modalities with individual layer normalizations (LNs) that individually compute the means and variances of different modalities. Specifically, parameters in the linear projection on patches, all linear projections ({\emph{e.g.}} for key, queries, etc) in MSA, and  MLP are shared for different modalities. Such a mechanism largely reduces the total model size which as discussed in the supplementary materials, even achieves better performance than using individual transformers. In addition, we also adopt shared positional embeddings for different modalities. We let the sparsity weight $\lambda=10^{-4}$ in \cref{eq:overall-loss} and the threshold $\theta=2\times10^{-2}$ in \cref{eq:mix-process-allocated} for all these experiments.

Our evaluation metrics include FID/KID for RGB predictions and MAE/MSE for other predictions. These metrics are introduced in the supplementary materials.

\textbf{Results.} In Table~\ref{tabs:multimodal-pix2pix}, we provide comparisons with extensive baseline methods and a SOTA method~\cite{DBLP:conf/nips/WangHSXRH20} with the same data settings. All methods adopt the learned ensemble over the two predictions which are corresponded to the two modality branches. In addition, all predictions have the same resolution $256\times 256$ for a fair comparison.
Since most existing methods are based on CNNs, we further provide two baselines for transformer-based models including the baseline without feature fusion (only uses ensemble for the late fusion) and the feature fusion method. By comparison, our TokenFusion surpasses all the other methods with large margins. For example, in the Shade+Texture$\to$RGB task, our TokenFusion (S) achieves $43.92/0.94$ FID/KID scores, remarkably better than the current SOTA method CEN~\cite{DBLP:conf/nips/WangHSXRH20} with 29.8\% relative FID metric decrease. 

In supplementary materials, we consider more modality inputs up to 4 which evaluates our group allocation strategy.

\begin{figure*}[t!]
\centering
\includegraphics[scale=0.16]{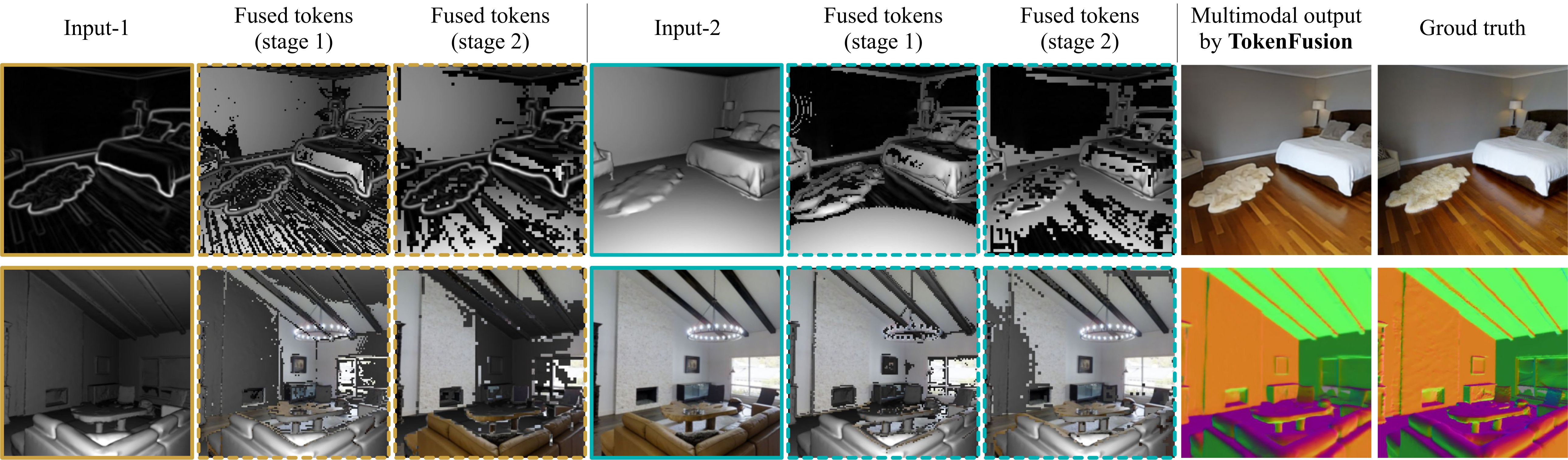}
\caption[]{Illustrations of which tokens are fused in our TokenFusion, performed on the \emph{validation} data split. We provide two cases including Texture+Shade$\to$RGB (first row) and Shade+RGB$\to$Normal (second row). The resolution of all images is $256\times256$. We choose the last layers in the first and second transformer stages respectively. {Best view in color and zoom in.}
}
\label{pic:pix2pix-mixt}\vskip-0.07in
\end{figure*}

\textbf{Visualization and analysis.} We provide qualitative results in \cref{pic:pix2pix-compare}, where we choose tough samples for comparison. The predictions with our TokenFusion obtain better natural patterns and are also richer in colors and details. In \cref{pic:pix2pix-mixt}, we further visualize the process of TokenFusion of which tokens are learned to be fused under our $l_1$ sparsity constraints. We observe that  the tokens for fusion follow specific regularities. For example, the texture modality tends to preserve its advantage of detailed boundaries, and meanwhile seek facial tokens from the shade modality. In this sense, TokenFusion combines complementary properties of different modalities.

\begin{table}[t]
	\centering
	\tablestyle{0.5pt}{1.05}
	\resizebox{1\linewidth}{!}{
		\begin{tabular}{l||c|c|c|c|c}
			\hskip0.02in Method
			& \fontsize{7pt}{1.2em}\selectfont \makecell[c]{Shade+Texture\\$\to$RGB$\;$}
			& \fontsize{7pt}{1.2em}\selectfont \makecell[c]{Depth+Normal\\$\to$RGB$\;$}
			& \fontsize{7pt}{1.2em}\selectfont \makecell[c]{RGB+Shade\\$\to$Normal$\,$}
			& \fontsize{7pt}{1.2em}\selectfont \makecell[c]{RGB+Normal\\$\to$Shade$\;$}
			& \fontsize{7pt}{1.2em}\selectfont \makecell[c]{RGB+Edge\\$\to$Depth$\;$}\\
			\shline
			\multicolumn{6}{c}{ CNN-based models}\\
			\cdashline{1-6}[1pt/1pt]
			Concat~\cite{DBLP:conf/nips/WangHSXRH20}&78.82$/$3.13 & $\;\,$99.08$/$4.28 & 1.34$/$2.85 & 1.28$/$2.02 &0.33$/$0.75 \\
			Self-Att.~\cite{DBLP:conf/nips/WangHSXRH20,journals/ijcv/ValadaMB20}&73.87$/$2.46 & $\;\,$96.73$/$3.95 & 1.26$/$2.76 & 1.18$/$1.76 &0.30$/$0.70 \\
			Align.~\cite{DBLP:conf/nips/WangHSXRH20,journals/tip/SongLLG20}& 92.30$/$4.20 & 105.03$/$4.91  & 1.52$/$3.25 & 1.41$/$2.21 & 0.45$/$0.90 \\
			CEN~\cite{DBLP:conf/nips/WangHSXRH20}& 62.63$/$1.65 & $\;\,$84.33$/$2.70 &1.12$/$2.51 & 1.10$/$1.72 & 0.28$/$0.66 \\
			\hline
			\multicolumn{6}{c}{Transformer-based models}\\
			\cdashline{1-6}[1pt/1pt]
			Concat (Ti)& \color{black}{76.13$/$2.85} & \color{black}{102.70$/$4.74}  & \color{black}{1.52$/$3.15} & \color{black}{1.33$/$2.20} & \color{black}{0.40$/$0.83}\\
			\rowcolor{gray!15}
			Ours (Ti)& 50.40$/$1.03 & $\;\,$76.35$/$2.19  & 0.73$/$1.83 & 0.95$/$1.54 & 0.21$/$0.57\\
			Concat (S)&\color{black}{72.55$/$2.39} & \color{black}{$\;\,$96.04$/$4.09} & \color{black}{1.18$/$2.73}&\color{black}{1.30$/$2.07}&\color{black}{0.35$/$0.68}\\
			\rowcolor{gray!15}
			Ours (S)&\textbf{43.92}$/$\textbf{0.94} & $\;\,$\textbf{70.13}$/$\textbf{1.92} &\textbf{0.58}$/$\textbf{1.51}&\textbf{0.79}$/$\textbf{1.33}&\textbf{0.16}$/$\textbf{0.47}\\	
		\end{tabular}
	}
\vskip-0.06in
\caption{Results on Taskonomy for multimodal image-to-image translation. Evaluation metrics are FID$/$KID ($\times 10^{-2}$) for RGB predictions and MAE ($\times 10^{-1}$)$/$MSE ($\times 10^{-1}$) for other predictions. Lower values indicate better performance for all the metrics. } 
\label{tabs:multimodal-pix2pix}
\vskip-0.17in
\end{table}

\begin{figure*}[t!]
\centering
\includegraphics[scale=0.2]{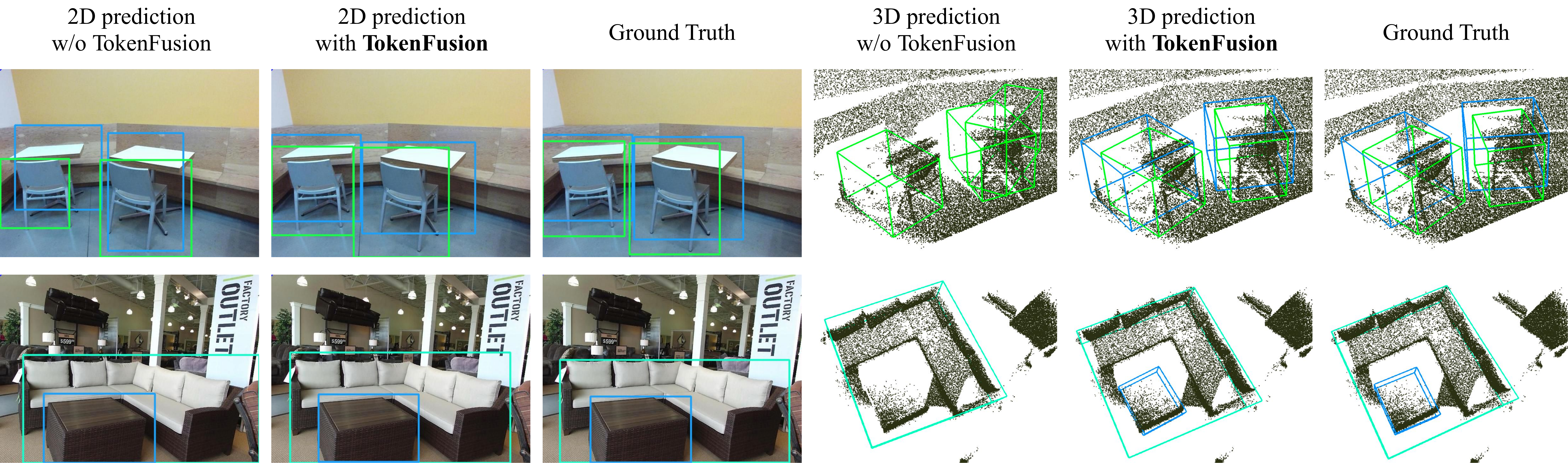}
\caption[]{Results visualization on the \emph{validation} data split for heterogeneous modalities including   point clouds and  images, where 3D object detection and 2D object detection are learned simultaneously. We compare the performance without (w/o) or with our TokenFusion. Our TokenFusion mainly benefits 3D object detection results.
}
\label{pic:3d-detection-compare}
\vskip-0.1in
\end{figure*}

\subsection{RGB-Depth Semantic Segmentation} 
\label{subsec:exp-segmentation}
We then evaluate  TokenFusion on another homogeneous  scenario, semantic segmentation with RGB and depth as input, which is a very common multimodal task and numerous methods have been proposed towards better performance. We choose the typical indoor datasets, NYUDv2~\cite{conf/eccv/SilbermanHKF12} and SUN RGB-D~\cite{conf/cvpr/SongLX15}. For NYUDv2, we follow the standard 795/654 images for train/test splits to predict the standard 40 classes~\cite{conf/cvpr/GuptaAM13}. SUN RGB-D is one of the most challenging large-scale indoor datasets, and we adopt the standard  5,285/5,050 images for train/test of 37 semantic classes. 

Our models include TokenFusion (tiny) and TokenFusion (small), of which the single-modal backbones follow  B2 and B3 settings of SegFormer~\cite{xie2021segformer}. Both tiny and small versions adopt the pretrained parameters on ImageNet-1$k$ for initialization following ~\cite{xie2021segformer}. Similar to our implementation in \cref{subsec:exp-img2img}, we also adopt shared transformers and positional embeddings for RGB and depth inputs  with individual LNs. We let the sparsity weight $\lambda=10^{-3}$ in \cref{eq:overall-loss} and the threshold $\theta=2\times10^{-2}$ in \cref{eq:mix-process-allocated} for all these experiments. 

\textbf{Results.} Results provided in Table~\ref{table:seg} conclude that current transformer-based models equipped with our TokenFusion surpass SOTA models using CNNs. Note that we choose relatively light backbone settings (B1 and B2 as mentioned in \cref{subsec:exp-segmentation}). We expect that using larger backbones (\eg, B5) would yield better performance.

\begin{table}[t]
	\centering
	 \vskip 0.05in
	\tablestyle{1.7pt}{0.96}
	\resizebox{1\linewidth}{!}{
		\begin{tabular}{l|c||ccc|cccc}
&&\multicolumn{3}{c|}{NYUDv2}&\multicolumn{3}{c}{SUN RGB-D}\\
\multirow{-2}*{Method}&\multirow{-2}*{Inputs}&\makecell[l]{Pixel Acc.} & \makecell[l]{mAcc.} & \makecell[l]{mIoU}& \makecell[l]{Pixel Acc.} & \makecell[l]{mAcc.} & \makecell[l]{mIoU} \\
			\shline
			\multicolumn{8}{c}{CNN-based models}\\
			\cdashline{1-8}[1pt/1pt]
			FCN-32s  \cite{journals/corr/LongSD14}&RGB&60.0&42.2&29.2&68.4&41.1&29.0\\
			RefineNet \cite{lin2019refinenet}&RGB&74.4&59.6&47.6&81.1&57.7&47.0\\
			FuseNet  \cite{conf/accv/HazirbasMDC16}&RGB+D  & 68.1&50.4&37.9&76.3&48.3&37.3\\
SSMA \cite{journals/ijcv/ValadaMB20}&RGB+D &75.2&60.5&48.7&81.0&58.1&45.7\\
RDFNet \cite{conf/iccv/LeePH17}&RGB+D   &76.0&62.8&50.1&81.5&60.1&47.7\\
AsymFusion~\cite{wang2020asymfusion}&RGB+D  &{77.0}&{64.0}&{51.2}&{-}&{-}&{-}\\
CEN \cite{DBLP:conf/nips/WangHSXRH20}&RGB+D  &{77.7}&{65.0}&{52.5}&{83.5}&{63.2}&{51.1}\\
			\hline
			\multicolumn{8}{c}{Transformer-based models}\\
			\cdashline{1-8}[1pt/1pt]
			w/o fusion (Ti)&RGB &75.2 & 62.5& 49.7 &82.3 & 60.6& 47.0\\
			Concat (Ti)&RGB+D&  76.5& 63.4&50.8 & 82.8& 61.4& 47.9\\
			\rowcolor{gray!15}
			Ours (Ti)&RGB+D  & 78.6&66.2 & 53.3& 84.0 & 63.3& 51.4\\
			w/o fusion (S)&RGB  &76.0 &63.0 &50.6&82.9 & 61.3&48.1\\
			Concat (S)&RGB+D  & 77.1 & 63.8& 51.4&83.5 & 62.0&49.0\\
			\rowcolor{gray!15}
			Ours (S)&RGB+D  & \textbf{79.0} & \textbf{66.9} &\textbf{54.2}& \textbf{84.7} & \textbf{64.1} &\textbf{53.0}\\
		\end{tabular}
	}\vskip-0.03in
\caption{Comparison results on the NYUDv2 and SUN RGB-D datasets with SOTAs for RGB and depth (D) semantic segmentation. Evaluation metrics include pixel accuracy (Pixel Acc.) (\%),  mean accuracy (mAcc.) (\%), and mean IoU (mIoU) (\%).}
\label{table:seg}
\vskip-0.1in
\end{table}

\subsection{Vision and Point Cloud 3D Object Detection} 
We further apply TokenFusion for fusing heterogeneous modalities, specifically, the 3D object detection task which has  received great attention. We leverage 3D point clouds and 2D images to learn 3D and 2D  detections, respectively, and both processes are learned simultaneously. We expect the involvement of 2D learning boosts the 3D counterpart.

We adopt SUN RGB-D~\cite{song2015sun} and ScanNetV2~\cite{DBLP:conf/cvpr/DaiCSHFN17} datasets. For SUN RGB-D, we follow the same train/test splits as in \cref{subsec:exp-segmentation} and detect the 10 most common classes. For ScanNetV2, we adopt the 1,201/312 scans as train/test splits to detect the 18 object classes. All these settings (splits and detected target classes) follow current works~\cite{qi2019deep,liu2021group} for a fair comparison. Note that different from SUN RGB-D, ScanNetV2 provides multi-view images for each scene alongside the point cloud. We randomly sample 10 frames per scene from the scannet-frames-25k samples provided in~\cite{DBLP:conf/cvpr/DaiCSHFN17}.

Our architectures for 3D detection and 2D detection follow  GF~\cite{liu2021group} and YOLOS~\cite{fang2021you}, respectively. We adopt the ``L6, O256'' or ``L12, O512'' versions of GF for the 3D  detection branch. We combine GF with the tiny (Ti) and small (S) versions of YOLOS, respectively, and adopt mAP@0.25 and mAP@0.5 as evaluation metrics following~\cite{qi2019deep,liu2021group}.

\textbf{Results.}
\label{sec:detection-results}
 We provide results comparison in Table~\ref{table:sunrgbd-detection} and Table~\ref{table:scannet-detection}. The main comparison is based on the best results of five experiments between different methods, and numbers within the brackets are the average results. Besides, we perform intuitive multimodal experiments by appending the 3-channel RGB vectors to the sampled points after PointNet++~\cite{DBLP:conf/nips/QiYSG17}. Such intuitive experiments are marked by the subscript * in both tables. We observe, however, that simply appending RGB information even leads to the performance drop, indicating the difficulty of such a heterogeneous fusion task. By comparison, our TokenFusion achieves new records on both datasets, which are remarkably superior to previous CNN/transformer  models in terms of both metrics. For example, with  TokenFusion, YOLOS-Ti can be utilized to boost the performance of GF by further 2.4 mAP@0.25 improvements, and using YOLOS-S  brings further gains.
 
 \textbf{Visualizations.} \cref{pic:3d-detection-compare} illustrates the comparison of detection results when using  TokenFusion for multimodal interactions against individual learning. We observe that TokenFusion  benefits the 3D detection part. For example, with the help of images, models with TokenFusion can locate 3D objects even with sparse or missing point data (second row). In addition, using images also benefits when the points of two objects are largely overlapped (first row). These observations demonstrate the advantages of our TokenFusion.

\begin{table}[t]
	\centering
	\vskip -0.02in
	\tablestyle{0.6pt}{1}
	\resizebox{1\linewidth}{!}{
		\begin{tabular}{l|c|c||c|c}
			\hskip0.02in Method
			& Backbone
			& Inputs
			& mAP@0.25
			& mAP@0.5\\
			\shline
			\multicolumn{5}{c}{CNN-based models}\\
			\cdashline{1-5}[1pt/1pt]
			VoteNet~\cite{qi2019votenet}&PointNet++ & Points & 59.1 & 35.8 \\
			VoteNet~\cite{qi2019votenet}*&PointNet++ & Points+RGB & 58.0 & 34.3 \\
			MLCVNet~\cite{xie2020mlcvnet}&PointNet++ & Points & 59.8 & -  \\
			HGNet~\cite{chen2020hierarchical}&GU-net & Points & 60.1 & 39.0 \\
			H3DNet~\cite{zhang2020h3dnet}& $4\times$PointNet++ & Points & 61.6 & -\\
			imVoteNet~\cite{qi2020imvotenet}& PointNet++ & Points+RGB & 63.4 & -\\
			\hline
			\multicolumn{5}{c}{Transformer-based models}\\
			\cdashline{1-5}[1pt/1pt]
			\makecell[l]{GF~\cite{liu2021group} (L6, O256)}& PointNet++ & Points & 63.0 (62.6) & 45.2 (44.4) \\
			\makecell[l]{GF~\cite{liu2021group} (L6, O256)*}& PointNet++ & Points+RGB & 62.1 (61.0) & 42.7 (41.9) \\
			\rowcolor{gray!15}
			\makecell[l]{Ours (L6, O256; Ti)$\;$} &PointNet++ & Points+RGB & {64.5 (64.2)} & {47.8 (47.3)} \\
			\rowcolor{gray!15}
			\makecell[l]{Ours (L6, O256; S)$\;\;$} &PointNet++ & Points+RGB & \textbf{64.9} (64.4) & \textbf{48.3} (47.7) \\
		\end{tabular}
	}\vskip-0.03in
\caption{Comparison on SUN RGB-D with SOTAs for 3D object detection, including best results and average results in brackets.  * indicates appending RGB  to the points as described in \cref{sec:detection-results}.}\vskip-0.1in
\label{table:sunrgbd-detection}

\end{table}

\begin{table}[t]
	\centering
	
	\tablestyle{0.5pt}{1}
	\resizebox{1\linewidth}{!}{
		\begin{tabular}{l|c|c||c|c}
			\hskip0.02in Method
			& Backbone
			& Inputs
			& mAP@0.25
			& mAP@0.5\\
			\shline
			\multicolumn{5}{c}{CNN-based models}\\
			\cdashline{1-5}[1pt/1pt]
			HGNet~\cite{chen2020hierarchical}&GU-net & Points& 61.3 & 34.4 \\
			GSDN~\cite{gwak2020generative} & MinkNet & Points& 62.8 & 34.8 \\
			3D-MPA~\cite{engelmann20203d} & MinkNet& Points & 64.2 & 49.2 \\
			VoteNet~\cite{qi2019votenet} & PointNet++ & Points&62.9 & 39.9 \\
			MLCVNet~\cite{xie2020mlcvnet} & PointNet++ & Points& 64.5 & 41.4 \\
			H3DNet~\cite{zhang2020h3dnet} & PointNet++ & Points& 64.4 & 43.4 \\
			H3DNet~\cite{zhang2020h3dnet} & 4$\times$PointNet++ & Points& 67.2 & 48.1 \\
			\hline
			\multicolumn{5}{c}{Transformer-based models}\\
			\cdashline{1-5}[1pt/1pt]
			\makecell[l]{GF~\cite{liu2021group} (L6, O256)}& PointNet++ & Points & 67.3 (66.3)& 48.9 (48.5)  \\
			\makecell[l]{GF~\cite{liu2021group} (L6, O256)*}& PointNet++ & Points+RGB & 66.3 (65.7) & 47.5 (47.0) \\
			\makecell[l]{GF~\cite{liu2021group} (L12, O512)}& PointNet++w2$\times$ & Points & {69.1} (68.6) & {52.8} (51.8)\\
			\makecell[l]{GF~\cite{liu2021group} (L12, O512)*}& PointNet++w2$\times$ & Points+RGB & 68.2 (67.6) & 50.3 (49.4) \\
			\rowcolor{gray!15}
			\makecell[l]{Ours (L6, O256; Ti)$\;\;\;$} &PointNet++ & Points+RGB & 68.8 (68.0) & 51.9 (51.2) \\
			\rowcolor{gray!15}
			\makecell[l]{Ours (L12, O512; S)$\;\;$} &PointNet++w2$\times$ & Points+RGB &\textbf{70.8} (69.8) & \textbf{54.2} (53.6) \\
		\end{tabular}
	}\vskip -0.01in
\caption{Comparison  on ScanNetV2 with SOTAs for 3D object detection, including best results and average results in brackets. }
\label{table:scannet-detection}\vskip-0.05in\vskip-0.03in

\end{table}

\begin{table}[t]
	\centering
	 \vskip -0.02in
	\tablestyle{1.7pt}{1}
	\resizebox{1\linewidth}{!}{
		\begin{tabular}{c|c||ccc|cccc}
&&\multicolumn{3}{c|}{Seg. (NYUDv2)}&\multicolumn{2}{c}{3D det.  (SUN RGB-D)}\\
\multirow{-2}*{$l_1$-norm}&\multirow{-2}*{\makecell[c]{Fusion strategy}}&\makecell[l]{Pixel Acc.} & \makecell[l]{mAcc.} & \makecell[l]{mIoU}& mAP@0.25 & mAP@0.5  \\
			\shline
			$\times$& $\times$&75.2 & 62.5& 49.7 &62.8 & 45.1\\
			$\times$& Random (10\%)&75.6 & 63.0& 50.1 &62.3 & 44.5\\
			$\times$& Random (30\%)&74.2 & 61.0& 48.2 &59.5 & 42.4\\
			\hline
			\checkmark& $\times$&75.0 & 62.5& 49.5 &62.6 & 44.9\\
			\rowcolor{gray!15}
			\checkmark&\checkmark (with RPA) & {78.6} & {66.2} &{53.3}& {64.9} & {48.3} \\
		\end{tabular}
	}
\caption{Effectiveness of $l_1$-norm and token fusion. Experiments include RGB-depth segmentation (seg.) on NYUDv2 and 3D detection (det.) with images and points on SUN RGB-D.}
\label{table:ablation-l1-token}\vskip-0.02in
\end{table}

\begin{table}[t]
	\centering
	\tablestyle{3pt}{1}
	\resizebox{1\linewidth}{!}{
		\begin{tabular}{c|c||ccc|cccc}
&&\multicolumn{3}{c|}{Seg. (NYUDv2)}&\multicolumn{2}{c}{3D det. (SUN RGB-D)}\\
\multirow{-2}*{\makecell[c]{Token fusion\\ (with $l_1$-norm)}}&\multirow{-2}*{\makecell[c]{RPA}}&\makecell[l]{Pixel Acc.} & \makecell[l]{mAcc.} & \makecell[l]{mIoU}& mAP@0.25 & mAP@0.5  \\
			\shline
			$\times$& $\times$&75.2 & 62.5& 49.7 &62.8 & 45.1\\
			$\times$& \checkmark&75.7 & 62.9& 50.3 &63.0 & 45.3\\
			\hline
			\checkmark& $\times$&78.3 & 65.8& 52.9 &63.6 & 46.2\\
			\rowcolor{gray!15}
			\checkmark&\checkmark  & {78.6} & {66.2} &{53.3}& {64.9} & {48.3} \\
		\end{tabular}
	}
\caption{Effectiveness of RPA proposed in \cref{subsec:homogeneous-modalities}. Experimental tasks and datasets follow Table~\ref{table:ablation-l1-token}.}
\label{table:ablation-rpa}\vskip-0.03in
\end{table}

\section{Ablation Study}
\textbf{$l_1$-norm and token fusion.} In Table~\ref{table:ablation-l1-token}, we demonstrate the advantages of $l_1$-norm and token fusion. We additionally conduct experiments with random token fusion. We observe that applying $l_1$-norm itself has little effect on the performance yet it is essential to reveal tokens for fusion. Our token fusion together with $l_1$-norm achieves much better performance than the random fusion baselines.

\textbf{Evaluation of RPA.} Table~\ref{table:ablation-rpa} evaluates  RPA proposed in \cref{subsec:rpa}. Results indicate that only using RPA without token fusion does not noticeably affect the performance, but is important when combined with the token fusion process for alignments, especially for the  3D detection task.

\section{Conclusion}
We propose TokenFusion, an adaptive method generally applicable for fusing  vision transformers with homogeneous or heterogeneous modalities. TokenFusion exploits uninformative tokens and re-utilizes these tokens to strengthen the interaction of other informative multimodal tokens. Alignment relations of different modalities can be explicitly utilized due to our residual positional alignment and inter-modal projection. TokenFusion surpasses state-of-the-art methods on  a variety of tasks, demonstrating its superiority and generality for multimodal fusion.

\section*{Acknowledgement}
This work is funded by Major Project of the New Generation of Artificial Intelligence (No. 2018AAA0102900) and the Sino-German  Collaborative Research Project Crossmodal Learning (NSFC  62061136001/DFG TRR169). We gratefully acknowledge the support of MindSpore, CANN and Ascend AI Processor used for this research.

\clearpage 
\section*{\LARGE Appendix}
\appendix

\section{Additional Results}
\label{sec:results}

\textbf{Multiple input modalities.} In Table~\ref{tabs:pix2pix-more-modal}, we further evaluate our TokenFusion with more modality inputs from 1 to 4. When the number of input modalities is larger than 2, we adopt the group allocation strategy as proposed in Sec. 3.4 of our main paper. By comparison, the performance is consistently improved  when using more modalities, and TokenFusion is again noticeably better than CEN~\cite{DBLP:conf/nips/WangHSXRH20}, suggesting the ability to absorb information from more modalities. 

\begin{table}[h]
\centering
\tablestyle{4pt}{1.1}
	\resizebox{1\linewidth}{!}{
\begin{tabular}{l||c>{\columncolor{gray!12} }c>{\columncolor{gray!12} }cc|ccc}
Modality&CEN~\cite{DBLP:conf/nips/WangHSXRH20}&Ours (Ti)&Ours (S)\\
\shline
Depth&113.91$/$5.68 &108.16$/$5.50&$\;\;$97.13$/$4.97\\
Normal& 108.20$/$5.42&112.25$/$5.77&100.29$/$5.02\\
Texture& $\;\;$97.51$/$4.82&$\;\;$99.70$/$5.14&$\;\;$94.92$/$4.38\\
Shade& 100.96$/$5.17&104.73$/$5.43&$\;\;$97.35$/$4.77\\
\hline
\makecell[l]{Depth+Normal}& $\;\;$84.33$/$2.70&$\;\;$71.82$/$2.36&$\;\;${64.20}$/${1.69}\\
\makecell[l]{Depth+Normal+Texture}& $\;\;$60.90$/$1.56&$\;\;$53.17$/$1.22&$\;\;${42.54}$/${0.93}\\
\makecell[l]{Depth+Normal+Texture+Shade}& $\;\;$57.19$/$1.33&$\;\;$47.69$/$1.01&$\;\;${39.15}$/${0.81}\\
\end{tabular}}
\caption{Results on the Taskonomy dataset for multimodal image-to-image translation (to RGB) with $1\sim4$ modalities.}
\label{tabs:pix2pix-more-modal}
\end{table}

\textbf{Network sharing.} As mentioned in Sec. 3.4 of our main paper, we adopt shared parameters in both Multi-head Self-Attention (MSA) and Multi-Layer Perception (MLP) for  the fusion with homogeneous modalities, and rely on modality-specific Layer Normalization (LN) layers to uncouple the normalization process. Such network sharing technique is evaluated by our experiments including multimodal image-to-image translation (in Sec. 4.1) and RGB-depth semantic segmentation (in Sec. 4.2), which largely reduces the model size, and also enables the reuse of attention weights for different modalities. In Table~\ref{table:sharing-scheme}, we further conduct ablation studies to demonstrate the effectiveness of our network sharing scheme. Fortunately, the comparison indicates that our default setting (\ie, Shared MSA and MLP, individual LN) achieves a win-win scenario: apart from the advantage on storage efficiency, also achieves better results than using individual MSA and MLP on both tasks. Note that further sharing LN layers leads to the performance drop, especially on the image-to-image translation task. In addition, we adopt shared Positional Embeddings (PEs) by default for the fusion with homogeneous modalities, and we observe that sharing/unsharing PEs can achieve comparable performance in practice.

\begin{table}[t]
	\centering
	\tablestyle{3pt}{1}
	\resizebox{1\linewidth}{!}{
		\begin{tabular}{c|c||cc|ccccc}
&&\multicolumn{2}{c|}{Image translation}&\multicolumn{3}{c}{Seg. (NYUDv2)}\\
\multirow{-2}*{\makecell[c]{MSA\&MLP}}&\multirow{-2}*{\makecell[c]{LN}}&FID&KID {($\times 10^{-2}$)}
&\makecell[l]{Pixel Acc.} & \makecell[l]{mAcc.} & \makecell[l]{mIoU}\\
			\shline
			Unshared&Unshared&49.73 & 1.06& 78.3 &65.6 & 52.9\\
			Shared& Shared&67.45 & 1.82& 76.7 &63.8 & 52.0\\
			\rowcolor{gray!15}
			Shared& Unshared&43.92 & 0.94& 78.6 &66.2 & 53.3\\
		\end{tabular}
	}
\caption{Results comparison when using different network sharing schemes for image-to-image translation (Shade+Texture$\to$RGB) on Taskonomy and RGB-depth segmentation (seg.) on NYUDv2. Lower FID or KID values indicate better performance. }
\label{table:sharing-scheme}\vskip-0.01in
\end{table}

\textbf{Combining TokenFusion with channel-wise fusion.} Our TokenFusion  detects uninformative tokens and re-utilizes these tokens for multimodal fusion. We may further combine TokenFusion with  an orthogonal  method by channel-wise pruning which automatically detects uninformative channels. Different from the token-wise fusion method in TokenFusion, the channel-wise fusion is not conditional on input features. Inspired by CEN~\cite{DBLP:conf/nips/WangHSXRH20},  we leverage the scaling factors $\gamma$ of layer normalization (LN) to perform channel-wise pruning, and apply sparsity constraints on $\gamma$. LN in transformers performs normalization on its input $\bm{x}_{m,l}$.

\begin{figure*}[t!]
\centering
\includegraphics[scale=0.16]{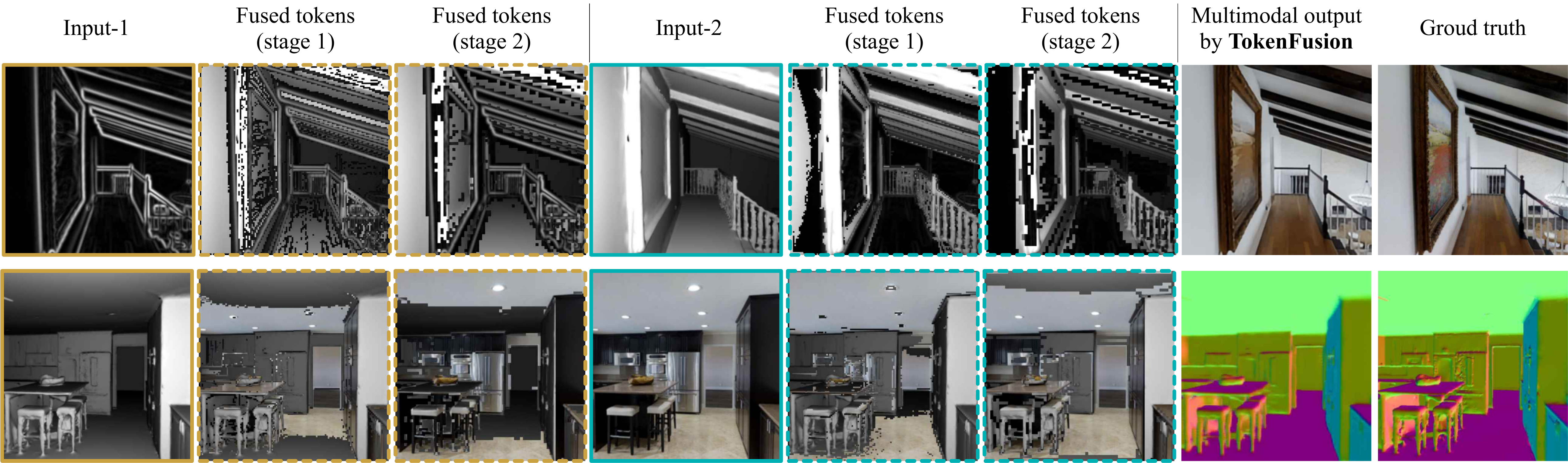}
\caption[]{Additional illustrations of the token fusion process as a supplement to Fig. 4 (main paper), performed on the \emph{validation} data split of Taskonomy. We provide two cases: Texture+Shade$\to$RGB (first row) and Shade+RGB$\to$Normal (second row). The resolution of all images is $256\times256$. We choose the last layers in the first and second transformer stages respectively. {Best view in color and zoom in.}
}
\label{pic:pix2pix-mixt2}\vskip-0.07in
\end{figure*}

To prune uninformative channels, we add a channel-wise pruning loss $\sum_{m=1}^M\sum_{l=1}^L|\bm{\gamma}_{m}^l|$ to the main loss in Eq. (5) (main paper). The overall loss function is 
\begin{equation}
\label{eq:overall-loss}
\mathcal{L}=\sum_{m=1}^M\Big(\mathcal{L}_m+\lambda_1\sum_{l=1}^L\big|s^l(\bm{e}_m^{l})\big|+\lambda_2\sum_{l=1}^L|\bm{\gamma}_{m}^l|\Big),
\end{equation}
where $\lambda_1,\lambda_2$ are hyper-parameters for balancing different losses; $\bm{\gamma}_{m}^l$ is a vector with the length $C$, representing the scaling factor of LN at the $l$-th layer of the $m$-th modality.

\begin{table}[t]
	\centering
	\tablestyle{8pt}{0.95}
	\resizebox{1\linewidth}{!}{
		\begin{tabular}{c|c||ccccccc}
&&\multicolumn{3}{c}{Seg. (NYUDv2)}\\
\multirow{-2}*{\makecell[c]{Token-wise}}&\multirow{-2}*{\makecell[c]{Channel-wise}}
&\makecell[l]{Pixel Acc.} & \makecell[l]{mAcc.} & \makecell[l]{mIoU}\\
			\shline
			$\times$&$\times$&75.2 & 62.5& 49.7 \\
			\rowcolor{gray!15}
			\checkmark&$\times$&78.6 & 66.2& 53.3 \\
			$\times$&\checkmark&77.2 & 65.0& 52.1\\
			\checkmark&\checkmark&78.8 & 66.6& 53.8 \\

		\end{tabular}
	}
\caption{RGB-depth segmentation results on the NYUDv2 dataset when combining our TokenFusion with the channel-wise fusion.}
\label{table:channel-wise}
\end{table}

\begin{table}[t]
	\centering
	\tablestyle{3.5pt}{1}
	\resizebox{1\linewidth}{!}{
		\begin{tabular}{c|c||cc||c}
&&\multicolumn{2}{c||}{3D det.  (ScanNetV2)}\\
\multirow{-2}*{\makecell[c]{Input image\\ frames}}& \multirow{-2}*{\makecell[c]{Model}}& mAP@0.25 & mAP@0.5 & \multirow{-2}*{\makecell[c]{Seconds per\\100 scenes}}\\
			\shline
			0&Ours (L6, O256; Ti)&67.3 & 49.0& 4.7 \\
			5&Ours (L6, O256; Ti)&67.9 & 50.5& 5.9 \\
			10&Ours (L6, O256; Ti)&68.8 & 51.9& 7.0 \\
		\end{tabular}
	}
\caption{Comparison of practical inference speed on ScanNetV2.}
\label{table:ablation-time}\vskip-0.07in
\end{table}

We  let $\lambda_1=\lambda_2=10^{-3}$ for RGB-depth segmentation experiments. 
Results provided in Table~\ref{table:channel-wise} demonstrate that our TokenFusion can be combined with the channel-wise fusion to obtain a further improved performance. For example, the segmentation on NYUDv2 with both token-wise and channel-wise fusion achieves an additional 0.5 mIoU gain than TokenFusion. More detailed studies of such combined framework, the relation between the overall pruning rate and fusion performance gain,  and the extension to fuse heterogeneous modalities are left to be the future works.

\textbf{Additional visualizations.}  In \cref{pic:pix2pix-mixt2}, we  provide another group of visualizations that depict the fused tokens under the $l_1$ sparsity constraints during training. We observe that fused tokens follow the  regularities mentioned in our main paper, \eg, the texture modality preserves its advantage at boundaries while seeking facial tokens from the shade modality.

\textbf{Inference speed.} In Table~\ref{table:ablation-time}, we test the real inference speed (single V100, 256G RAM) with different numbers of input frames for 3D detection. We observe that additional time costs are mild, which is partly because the added YOLOS-Ti is a light model (with only three multi-heads).

\section{More Details of Image Translation}
\label{sec:imple}
In this part, we discuss the implementation details for our  image-to-image translation task. Our implementation contains two transformers as the generator and discriminator respectively.  The resolution of the generator/discriminator input or the generator prediction is  $256\times256$. Specifically, the discriminator of our model is similar to~\cite{jiang2021transgan}, which adopts five stages with two layers for each, where the embedding dimensions and head numbers gradually double from $32$ to $512$ and from $1$ to $16$ respectively. The generator is composed of nine stages where the first five have the same configurations with the discriminator, and the last four stages have reverse configurations of its first four stages. 

We adopt four kinds of evaluation metrics including Mean Square Error (MSE), Mean Absolute Error (MAE), Fréchet-Inception-Distance (FID), and Kernel-Inception-Distance (KID). Here we briefly introduce FID and KID scores. FID, proposed by \cite{conf/nips/HeuselRUNH17}, contrasts the statistics of generated samples against real samples. The FID fits a Gaussian distribution to the hidden activations of InceptionNet for each compared image set and then computes the Fréchet distance (also known as the Wasserstein-2 distance) between those Gaussians. Lower FID is better, corresponding to generated images more similar to the real. KID, developed by~\cite{conf/iclr/BinkowskiSAG18}, is a metric similar to the FID but uses the squared Maximum-Mean-Discrepancy (MMD) between Inception representations with a polynomial kernel. Unlike FID, KID has a simple unbiased estimator, making it more reliable especially when there are much more inception features channels than image numbers. Lower KID indicates more visual similarity between real and generated images. Regarding our implementation of KID, the hidden representations are derived from the Inception-v3~\cite{DBLP:conf/cvpr/SzegedyVISW16} pool3 layer.

{\small
\bibliographystyle{ieee_fullname}
\bibliography{egbib}
}

\end{document}